\pdfoutput=1

\documentclass[11pt]{article}
\usepackage{listings}
\usepackage[dvipsnames]{xcolor}

\usepackage{acl}

\usepackage{times}
\usepackage{latexsym}

\usepackage[T1]{fontenc}

\usepackage[utf8]{inputenc}

\usepackage{microtype}

\usepackage{inconsolata}
\usepackage{bbm}
\usepackage{amsmath}
\usepackage{graphicx}
\usepackage{multirow}
\usepackage{multicol}
\usepackage{subfigure}
\usepackage{hyperref}

\title{Tram: A Token-level Retrieval-augmented Mechanism for Source Code Summarization}

\author{
Tong Ye\textsuperscript{1}, 
Lingfei Wu\textsuperscript{2}, 
Tengfei Ma\textsuperscript{3}, 
Xuhong Zhang\textsuperscript{1},  
Yangkai Du\textsuperscript{1}, \\
\bf{Peiyu Liu}\textsuperscript{1},
\bf{Shouling Ji}\textsuperscript{1},
\bf{Wenhai Wang}\textsuperscript{1\thanks{\quad Corresponding author.}}\\
        \textsuperscript{1}Zhejiang University; \textsuperscript{2}Anytime.AI; 
        \textsuperscript{3}Stony Brook University\\
        \texttt{\{{tongye,zhangxuhong,yangkaidu,liupeiyu,sji,zdzzlab}\}@zju.edu.cn} \\
        \texttt{lwu@anytime-ai.com, tengfei.ma@stonybrook.edu}
        }

\begin{document}
\maketitle
\begin{abstract}
Automatically generating human-readable text describing the functionality of a program is the intent of source code summarization. Although neural language models achieve significant performance in this field, they are limited by their inability to access external knowledge. To address this limitation, an emerging trend is combining neural models with external knowledge through retrieval methods. Previous methods have relied on the sentence-level retrieval paradigm on the encoder side. However, this paradigm is coarse-grained, noise-filled and cannot directly take advantage of the high-quality retrieved summary tokens on the decoder side. In this paper, we propose a fine-grained Token-level retrieval-augmented mechanism (Tram) on the decoder side rather than the encoder side to enhance the performance of neural models and produce more low-frequency tokens in generating summaries. Furthermore, to overcome the challenge of token-level retrieval in capturing contextual code semantics, we also propose integrating code semantics into individual summary tokens. The results of extensive experiments and human evaluation show that our token-level retrieval-augmented approach significantly improves performance and is more interpretable. 
\end{abstract}

\section{Introduction}
With software functions becoming more comprehensive and complex, it becomes a heavy burden for developers to understand software.
It has been reported that nearly 90\% \citep{wan2018improving} of effort is used for maintenance, and much of this effort is spent on understanding the maintenance task and related software source codes. Source code summary as a natural language is indispensable in software since humans can easily read and understand it, as shown in Table \ref{python_instance}. However, manually writing source code summaries is time-consuming and tedious. Besides, the source code summary is often outdated in continuous software iteration. Hence, automatically generating concise, human-readable source code summaries is critical and meaningful.

\begin{table}[ht]
\centering
\begin{tabular}{|c|}
\hline 
\begin{lstlisting}[   % 进行参数设置
 language=Python, % 设置语言
 basicstyle=\small,
 basicstyle=\ttfamily\tiny, % 设置字体族
 breaklines=true, % 自动换行
 keywordstyle=\bfseries\color{NavyBlue}, % 设置关键字为粗体，颜色为 NavyBlue
 morekeywords={}, % 设置更多的关键字，用逗号分隔
 emph={np, interval}, % 指定强调词，如果有多个，用逗号隔开
    emphstyle=\bfseries\color{Rhodamine}, % 强调词样式设置
    commentstyle=\itshape\color{black!50!white}, % 设置注释样式，斜体，浅灰色
    stringstyle=\bfseries\color{PineGreen!90!black}, % 设置字符串样式
    columns=flexible,
    % frame={topline, bottomline, rightline},
]
def cos(x):
    np = import module("numpy")
    if isinstance(x, (int, float)):
        return interval(np.sin(x))
    elif isinstance(x, interval):
        if (not(np.isifnite(x.start) and 
                np.isfinite(x.end))):
            return interval((-1), 1, is_valid=x.is_valid)
        (na, _) = divmod(x.start, (np.pi / 2.0))
        (nb, _) = divmod(x.end, (np.pi / 2.0))
        start = min(np.cos(x.start), np.cos(x.end))
        end = max(np.cos(x.start), np.cos(x.end))
        if ((nb - na) > 4):
            return interval((-1), 1, is_valid=x.is_valid)
        elif (na == nb):
            return interval(start, end, is_valid=x.is_valid)
        else:
            if ((na // 4) != (nb // 4)):
                end = 1
            if (((na - 2) // 4) != ((nb - 2) // 4)):
                start = -1
            return interval(start, end, is_valid=x.is_valid)
    else:
        raise NotImplementedError
\end{lstlisting} \\
\hline \hline
\textbf{Summary:} evaluates the \textcolor[RGB]{205,50,120}{\textbf{cos}} of an interval. \\
\textbf{\textit{Token-level retrieval results}} \\
\textbf{\textit{at the next generation step "\textcolor[RGB]{205,50,120}{\textbf{cos}}":}} \\
cos, tangent, sin, hyperbolic, $\cdots$ \\
\hline 
\end{tabular}
\caption{A sample of source code summarization.}
\label{python_instance}
\end{table}

With the development of language models and the linguistic nature of source code, researchers explored Seq2Seq architecture, such as recurrent neural networks to generate summaries \citep{iyer_etal_2016_summarizing, loyola_etal_2017_neural, Liang_Zhu_2018}. Soon afterward, transformer-based models \citep{ahmad_etal_2020_transformer, wu_etal_2021_code, gong2022source} were proposed, outperforming previous RNN-based models by a large margin.
Recently, many approaches have been proposed to leverage the structural properties of source code, such as Abstract Syntax Tree (AST) and Program Dependency Graph (PDG). Current structure-aware methods typically either fuse structural information in a hybrid manner \citep{hu2018deep, shido2019automatic, leclair, choi_etal_2021_learning, shi_etal_2021_cast}, or use a structured-guided way \citep{wu_etal_2021_code, son_etal_2022_boosting, gong2022source, guo-etal-2022-modelinghie, choi-etal-2023-blocsum}. Although these methods have shown promising results, they primarily focus on leveraging the information within the code to obtain richer code representation without fully utilizing the potential of the available human-written code-summary pairs.

In order to leverage external existing high-quality code and the corresponding summary instances, recent works \citep{zhang2020retrieval, li2021editsum, liu2021retrievalaugmented, parvez2021retrieval} have proposed a retrieval augmented approach. Their unified paradigm involves sentence-level retrieval, which uses text similarity metrics or code semantic similarity metrics to retrieve the most similar code snippet from a code repository for the given input code snippet. The retrieved code snippet and its corresponding summary are either directly concatenated with the input code snippet or semantically enhanced to augment the input code snippet on the encoder side. 

However, the granularity of sentence-level retrieval methods poses challenges. Specifically, they can erroneously retrieve and incorporate code snippets that, while syntactically similar, are semantically distinct or those that only bear partial semantic resemblance. The unintended noise introduced through such mismatches can adversely affect the generation performance, especially for low-frequency tokens. Moreover, code summarization is essentially a generative task, the decoder autoregressively generates the summary tokens. However, previous sentence-level retrieval-augmented methods neglect to fuse the retrieved information on the decoder side, only doing so on the encoder side, which will result in the utilization pattern being indirect and insufficient.

These limitations have inspired us to explore a more fine-grained and sufficient retrieval approach on the summary generation process. In order to achieve the purpose of retrieving semantic similar summary tokens on the decoder side, we first construct a datastore to store the summary tokens and corresponding representations through a pre-trained base model offline. 
Meanwhile, to overcome the challenge of not fully utilizing code semantics on the encoder side when retrieving on the decoder side, we intelligently fuse summary token representation with code token representation and AST node representation with attention weight. This approach fully considers contextual code semantics associated with summary tokens. Then, at each generation step, the fused summary token representation is used to retrieve the top-$K$ most similar tokens. As illustrated in Table \ref{python_instance}, the token-level retrieval results at the next token generation step \textit{``cos''} are \textit{``cos, tangent, sin, hyperbolic, $\cdots$''}. The retrieved top-$K$ tokens are expanded to a probability distribution, which we refer to as the retrieval-based distribution. The retrieval-based distribution is then fused with the vanilla distribution to form the final distribution. Additionally, our proposed token-level retrieval mechanism can be seamlessly integrated with existing sentence-level retrieval methods and code-related large pre-trained models.

To facilitate future research, we have made our code publicly available\footnote{\url{https://github.com/tongye98/SourceCodeSummary}}. Overall, the main contributions of this paper can be outlined as follows:

(1) We are the first to explore a Token-level retrieval-augmented mechanism (Tram) on the decoder side for source code summarization.

(2) Our proposed retrieval-augmented mechanism is orthogonal to existing improvements, such as better code representation, additional sentence-level retrieval approaches, and pre-trained models.

(3) Extensive experiments and human evaluation show that Tram significantly outperforms other baseline models, generates more low-frequency tokens and is more interpretable.

\section{Related Works}
\paragraph{Retrieval-based Source Code Summarization.}
\citet{liu2021retrievalaugmented} retrieved the most similar code snippet by text similarity metric to enrich target code structure information for getting a better code representation encoder. This retrieval method only carries out from the perspective of text similarity and neglects code semantic similarity in the retrieval phase. Besides, the summary corresponding to the retrieved code snippet is just a simple concatenation to the encoder. \citet{zhang2020retrieval, parvez2021retrieval} used a pre-trained encoder to obtain code semantic representation, which was used to retrieve similar code snippets. The former only uses similar code snippets and discards the corresponding summaries; the latter directly splice the retrieved code snippet and the corresponding summary behind the target code; both are also aimed at better code representation on the encoder side. Different from the above sentence-level retrieval methods, Tram performs token-level retrieval augmentation at each step of the decoder that generates the next token.

\paragraph{K-Nearest-Neighbor Machine Translation.}
Recently, non-parametric methods have been successfully applied to neural machine translation \citep{khandelwal2021nearest, jiang_etal_2021_learning, zheng_etal_2021_adaptive, zheng_etal_2021_non_parametric}. These approaches complement advanced NMT models with external memory to alleviate the performance degradation in domain adaption. Compared to these works, we have fully accounted for the code's inherent structure and have intelligently integrated code semantics into the retrieval process. Additionally, we demonstrate how Tram integrates with sentence-level retrieval methods.

\begin{figure}[t]
    \centering
    \includegraphics[scale=0.5]{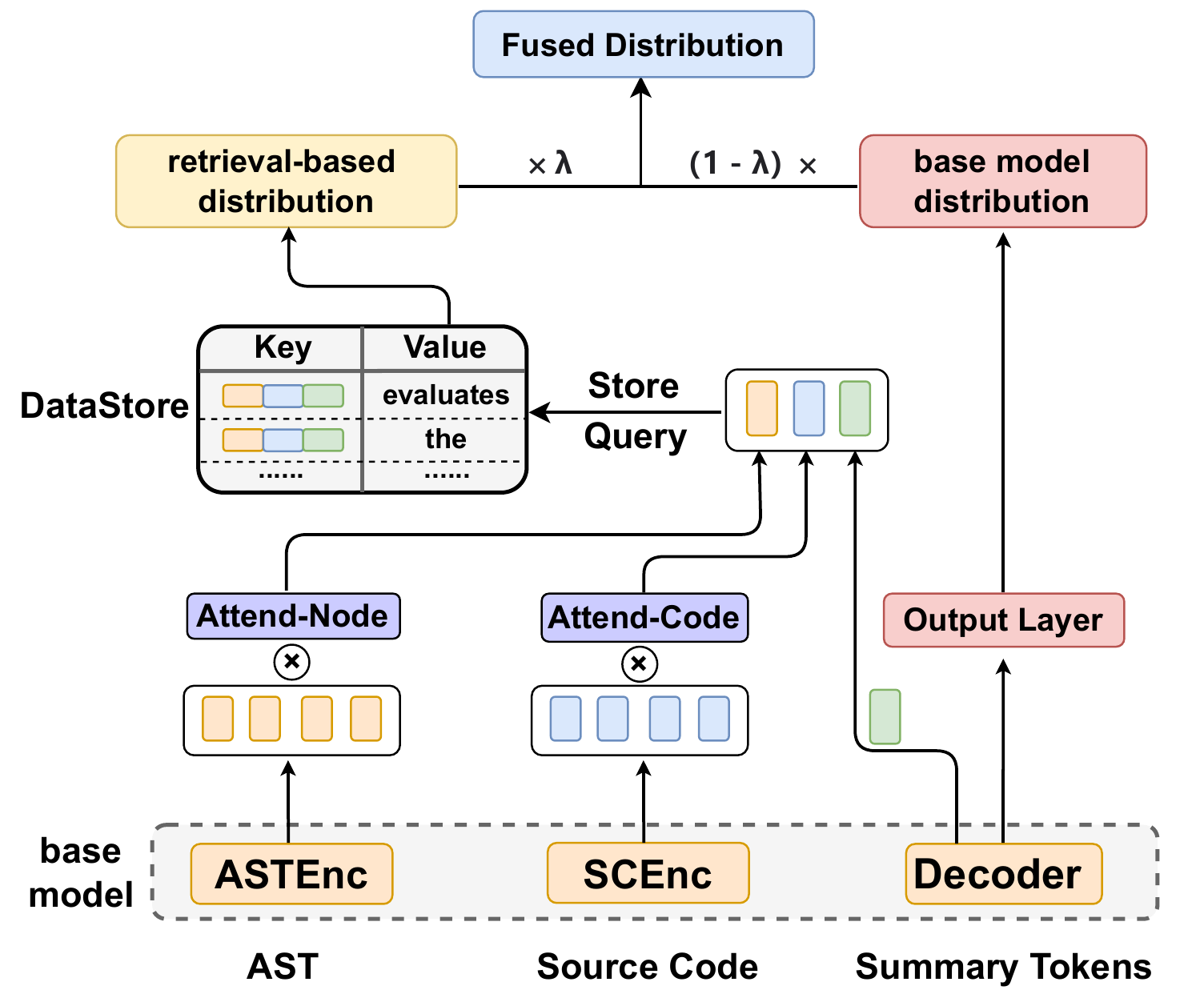}
    \caption{The overview architecture of Tram.}
    \label{fig:architecture}
\end{figure}

\section{Methodology}
\label{methodology}
\subsection{Overview}
The overview architecture of Tram is shown in Figure \ref{fig:architecture}. 
Initially, we introduce the base model, which is an encoder-decoder architecture that takes a code snippet and corresponding AST as input and generates a summary as output. Building upon the base model, we then construct a datastore that stores summary tokens and corresponding representations, where the representation is an intelligent combination of the decoder representation, code token representation, and AST node representation. Next, we develop a fine-grained token-level retrieval mechanism. This mechanism focuses on retrieving the top-$K$ most similar tokens from the datastore and generating a retrieval-based distribution. The retrieval-based distribution is then fused with the vanilla base model distribution by a weight hyper-parameter $\lambda$ to form the final distribution. Additionally, we detail the integration of both token-level and sentence-level retrieval. The combination of token-level retrieval and sentence-level retrieval enables a more comprehensive summarization process. In terms of integrating Tram with code pre-trained models, the implementation is broadly consistent and detailed in Appendix \ref{sec_tram_to_llms}.

\begin{figure}[t]
    \centering
    \includegraphics[scale=0.5]{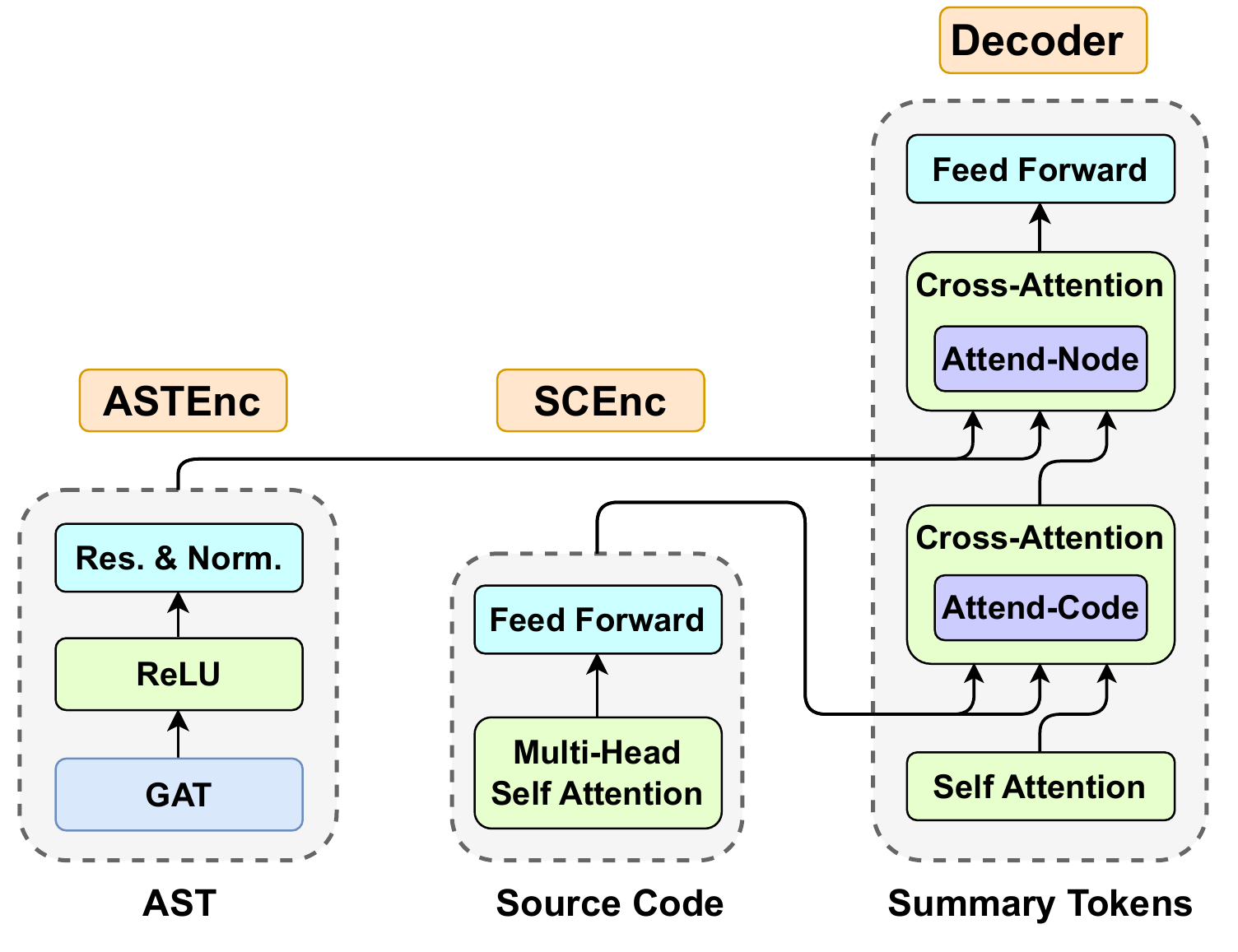}
    \caption{The architecture of base model.}
    \label{fig:base}
\end{figure}

\subsection{Base Model}
 The base model serves as the foundation for the subsequent retrieval process. It is designed to construct the datastore and generate the base model distribution. Figure \ref{fig:base} illustrates the specific architecture of the base model, which consists of two encoders (SCEnc and ASTEnc) and a decoder.
\paragraph{Source Code Encoder (SCEnc).}
As shown in Figure \ref{fig:base}, we utilize Transformer \citep{NIPS2017_3f5ee243} as the encoder for the source code tokens. The Transformer consists of stacked multi-head attention and parameterized linear transformation layers. Each layer emphasizes on self-attention mechanism. Nevertheless, as pointed out in \citet{ahmad_etal_2020_transformer}, the code semantic representation is influenced by the mutual interactions between its tokens rather than their absolute positions. Therefore, we adopt the method of relative positional encoding, as proposed by \citet{shaw_etal_2018_self}.

Assuming the code snippet contains $p$ tokens $[t_1,t_2,...,t_p]$, after SCEnc, each token has a hidden representation, which is denoted as:
\begin{equation}
[h_1,h_2,...,h_p] = SCEnc([t_1,t_2,...,t_p]) \nonumber
\end{equation} 

\paragraph{AST Encoder (ASTEnc).}
Furthermore, the AST of the source code can be considered as a graph structure, making it suitable for representation and learning using Graph Neural Networks (GNNs).
Taking advantage of the GAT's \citep{veličković2018graph} exceptional performance and its ability to assign adaptive attention weights to different nodes, we employ GAT to represent each node in the AST. The graph encoder layer processes the AST by first aggregating the neighbors of the nodes with edge information. It then updates the nodes with the aggregated information from their neighborhoods.

After updating the node information, the node representations are put together into a $ReLU$ activation followed by residual connection \citep{He_2016_CVPR} and layer normalization \citep{ba2016layer}.

Assuming the AST of the code snippet contains $q$ nodes $[n_1,n_2,...,n_q]$, after the ASTEnc, each node has a hidden representation, denoted as:
\begin{equation}
[r_1,r_2,...,r_q] = ASTEnc([n_1,n_2,...,n_q]) \nonumber
\end{equation} 

\paragraph{Summary Decoder.}
The summary decoder is designed with modified transformer decoding blocks. At time step $t$, given the existing summary tokens $[s_1,s_2,...,s_{t-1}]$, the decoding blocks first encode them by masked multi-head attention. After that, we expand the transformer block by leveraging two multi-head cross-attention modules to interact with the two encoders for summary decoding. One multi-head cross-attention module is performed over the code token features to get the first-stage decoded information, which will then be fed into the other over the learned AST node features for the second-stage decoding. Then the decoded summary vectors $[d_1,d_2,...,d_{t-1}]$ are put into a feed-forward network for non-linear transformation.

\subsection{Datastore Construction}
\label{datastore_construction}
Based on the base model, to achieve the goal of fine-grained token-level retrieval, we build the datastore that stores summary tokens and corresponding representations. At the stage of datastore establishment, we adopt the above pre-trained base model to go through all training instances in an offline manner. During this process, for each instance, the SCEnc and ASTEnc encode the code tokens and AST nodes into a sequence of hidden states: $[h_1,h_2,...,h_p]$ and $[r_1,r_2,...,r_q]$, the decoder generates the target summary autoregressively. At time step $t$, the decoder takes existing summary token $[s_1,s_2,...,s_{t-1}]$ as input, for the last token $s_{t-1}$, the decoder's first cross-attention module gets the attention score of the code tokens (called Attend-Code $[\alpha_1,\alpha_2,...,\alpha_p]$), the second cross-attention module gets the attention score of the AST nodes (called Attend-Node $[\beta_1,\beta_2,...\beta_q]$). We use Attend-Code and Attend-Node to perform weighted summation of the representations of code tokens and AST nodes, respectively, denoted as:
\begin{equation}
    [\alpha_1,\alpha_2,...,\alpha_p] * [h_1,h_2,...,h_p]^T = H_t \nonumber
\end{equation}
\begin{equation}
    [\beta_1,\beta_2,...,\beta_q] * [r_1,r_2,...,r_p]^T = R_t  \nonumber
\end{equation}
where $H_t$ means weighted code token representation, $R_t$ means weighted AST node representation.

After two cross-attention modules, the input token $s_{t-1}$ is converted to token representation $d_{t-1}$. Because the goal at time step $t$ is to generate the next token $s_t$, we pick the token representation $d_{t-1}$ to represent $s_t$. To fully consider the contextual code semantics associated with the summary token, we concatenate $H_t$, $R_t$, and $d_{t-1}$ to create the final and more comprehensive representation of $s_t$. Besides, to facilitate efficient retrieval in the subsequent steps, we applied $L_2$ regularization to the representations in practice, denoted as:
\begin{equation}
    k_t = Concat(H_t, R_t, d_{t-1}) \nonumber
\end{equation}
\begin{equation}
    \widetilde k_t = L_2\_Normalize(k_t) \nonumber
\end{equation}
where $\widetilde k_t$ is the final presentation of token $s_t$. Finally, the ground-truth summary token $s_t$ and corresponding representation $\widetilde k_t$ are inserted into datastore as a key-value pair, denoted as (key, value) = ($\widetilde k_t$, $s_t$), the whole datastore can be denoted as:
\begin{equation}
    (\mathcal{K}, \mathcal{V}) = \{ (\widetilde k_t, s_t), \forall s_t \in S \} \nonumber
\end{equation}
where $S$ means all summary tokens in the training dataset. It is important to note that the datastore contains duplicate tokens because the same summary token can have different keys, representing different semantic representations due to variations in linguistic contexts.

\subsection{Token-level Retrieval}
During inference, at each decoding step $t$, the current summary token representation $d_{t-1}$ is combined with the corresponding $H_t$ and $R_t$ using the same concatenate and $L_2$ regularization operator as query $q_t$. The query retrieves the top-$K$ most similar summary tokens in the datastore according to 
cosine similarity distance. It is worth noting that we use cosine similarity instead of squared-$L^2$ distance because of the performance of the preliminary experiment. As an added bonus, cosine similarity can be seen as retrieval confidence. In practice, the retrieval over millions of key-value pairs is carried out using FAISS \citep{johnson2019billion}, a library for fast nearest neighbor search in high-dimensional spaces. The retrieved key-value pairs $(k,v)$ and corresponding cosine similarity distance $\alpha$ composed a triple set $\mathcal{N} = \{ (k_i, v_i, \alpha_i) \vert i=1,2,\cdots, K\}$. Inspired by KNN-MT \citep{khandelwal2021nearest}, the triple set can then  be expanded and normalized to the retrieval-based distribution as follows:
\begin{equation}
    P_r(s_t | c, \hat{s}_{<t}) \propto \sum_{(k_i, v_i, \alpha_i) \in \mathcal{N}} \mathbbm{1}_{v_i=s_t} \exp \left( g(k_i, \alpha_i) \right) \nonumber
\end{equation}
\begin{equation}
    g(k_i, \alpha_i) = \alpha_i \ast T \nonumber
\end{equation}
where $g(\cdot)$ can be any Kernel Density Estimation (KDE); in practice, we use the product form; $T$ is the temperature to regulate probability distribution.

\subsection{Fused Distribution}
The final prediction distribution can be seen as a combination of the vanilla base model output distribution and the retrieval-based distribution, which is interpolated by a hyper-parameter $\lambda$:
\begin{equation}
    \begin{aligned}
    P(s_t | c, \hat{s}_{<t}) &= \lambda \ast P_r(s_t | c, \hat{s}_{<t}) \\
        &+ (1-\lambda) \ast P_m(s_t | c, \hat{s}_{<t}) \nonumber
    \end{aligned}
\end{equation}
where $P_m$ indicates the base model distribution.

\subsection{Additional Sentence-level Retrieval}
Our proposed token-level retrieval augmented method can also be seamlessly incorporated with additional sentence-level retrieval. 
Sentence-level retrieval here means using the target code snippet to retrieve the most semantically similar code snippet in the corpus through code semantic representations. Then we assign an additional but the same base model for the most similar code snippet to generate tokens autoregressively. At each generation step, the decoder of the additional base model (generating similar-code-based next token distribution ) is synchronous with the original target code snippet decoder (generating base model next token distribution). Finally, the above two distributions, together with the ``token-level retrieved next token distribution'', form the final distribution through a weighted sum, which is denoted as:

\begin{equation}
    \begin{aligned}
    P(s_t | c, \hat{s}_{<t}) &= \lambda_1 \ast P_r(s_t | c, \hat{s}_{<t}) \\ 
                             &+ \lambda_2 \ast Sim \ast  P_s(s_t | \langle c \rangle, \hat{s}_{<t})  \\
                             &+ (1-\lambda_1-\lambda_2) \ast P_m(s_t | c, \hat{s}_{<t}) \nonumber
    \end{aligned}
\end{equation}
where $P_s$ is the additional base model produced distribution, $\langle c \rangle$ is the most semantically similar code snippet to the target code snippet $c$, and $Sim$ is the corresponding similarity score.

\begin{table}[t]
    \centering
    \resizebox{\linewidth}{!}{
    \begin{tabular}{c|c|c|c|c}
    \hline
    \textbf{Datasets} & \textbf{Java} & \textbf{Python} &\textbf{CCSD} & \textbf{Python$^\ddag$}\\
    \hline
         Train &  69,708 & 55,538 & 84,316 & 65,236 \\
         Validation &  8,714 & 18,505 & 4,432 & 21,745\\
         Test & 8,714 & 18,502 & 4,203 & 21,745 \\
    \hline
    \small Code: Avg. tokens & 73.76 & 49.42 & 68.59 & 150.82\\
    \small Summary: Avg. tokens & 17.73 & 9.48 & 8.45 & 9.93\\
    \hline
    \end{tabular}
    }
    \caption{Statistics of the experimental datasets.}
    \label{tab: statistics}
\end{table}

\section{Experiments}
\subsection{Experimental Setup}
\paragraph{Datasets.}
We conduct the experiments on four public benchmarks of Java \citep{hu2018deep}, Python \citep{wan2018improving}, CCSD (C Code Summarization Dataset) \citep{liu2021retrievalaugmented}, and Python$^\ddag$ \citep{zhang2020retrieval}. The partitioning of train/validation/test sets follows the original datasets. The statistics of the four datasets are shown in Table \ref{tab: statistics}. 

\paragraph{Out-of-Vocabulary.} 
The vast operators and identifiers in program language may produce a much larger vocabulary than natural language, which can cause Out-of-Vocabulary problem. To avoid this problem, we apply $CamelCase$ and $snake_{-}case$ tokenizers that are consistent with recent works \citep{gong2022source, wu_etal_2021_code, ahmad_etal_2020_transformer} to reduce the vocabulary size of source code.

\begin{table*}[ht]
\centering
\resizebox{\textwidth}{!}{
    \begin{tabular}{l|ccc|ccc}
    \hline \hline
    \multirow{2}{*}{\textbf{Model}} & \multicolumn{3}{c}{\textbf{Java}} & \multicolumn{3}{|c}{\textbf{Python}} \\
    \cline{2-7}
    & \textbf{BLEU} & \textbf{ROUGE-L} & \textbf{METEOR} & \textbf{BLEU} & \textbf{ROUGE-L} & \textbf{METEOR} \\
    \hline
    \textit{Transformer-based Methods} & & & & & & \\
    Transformer \citep{ahmad_etal_2020_transformer} & 44.58 & 54.76 & 26.43 & 32.52 & 46.73 & 19.77 \\
    CAST \citep{shi_etal_2021_cast}                 & 45.19 & 55.08 & 27.88 & -     & -     & -     \\
    mAST + GCN \citep{choi_etal_2021_learning}      & 45.49 & 54.82 & 27.17 & 32.82 & 46.81 & 20.12 \\
    SiT \cite{wu_etal_2021_code}                    & 45.70 & 55.54 & 27.55 & 33.46 & 47.50 & 20.28 \\
    SiT + PDG \citep{son_etal_2022_boosting}       & 46.86 & 56.69 & -     & -     & -     & -     \\
    CODESCRIBE \citep{guo-etal-2022-modelinghie} & 46.93 & 56.18 & 29.13 & 34.44 & 49.02 & 20.91 \\
    \hline 
    \textit{Our Method}  & & & & & & \\
    \hspace{1em}Base                               & 46.84 & 56.92 & 28.71 & 34.20 & 48.37 & 20.99 \\
    \hspace{1em}Tram w/o HR                       & 47.85 & 57.51 & 29.28 & 35.37 & 49.31 & 21.53\\
    \hspace{1em}Tram              & \textbf{48.32}  & \textbf{58.13} & \textbf{29.56} & \textbf{35.97} & \textbf{49.92} & \textbf{22.09} \\
    \hspace{1em}\textit{Tram with SenRe}      & \textit{48.58} & \textit{58.43} & \textit{29.77} & \textit{36.23} & \textit{50.04} & \textit{22.23} \\
    \hline \hline 
    \textit{Our Method on Pre-trained Models}  & & & & & & \\
    \hspace{1em}CodeT5 \citep{wang-etal-2021-codet5}                          & 46.47 & 58.11 & 27.92 & 35.37 & 51.27 & 23.22 \\
    \hspace{1em}\textit{CodeT5 + Tram}                         & 47.85  & 59.32 & 28.75 & 36.23 & 52.08 & 24.13 \\
    \hspace{1em}UniXcoder \citep{guo-etal-2022-unixcoder}      & 45.32 & 56.61 & 26.52 & 35.89 & 51.17 & 23.11 \\
    \hspace{1em}\textit{UniXcoder + Tram}                      & 46.17  & 57.22 & 26.94 & 36.45 & 51.78 & 23.55 \\
    \hline \hline 
    \end{tabular}
}
\caption{\label{main_result_java_python}
Comparison of the performance of our method with other baseline methods on Java and Python benchmarks in terms of BLEU, ROUGE-L, and METEOR. The results of baseline models are reported in their original papers. `-' refers to no corresponding value from the paper. HR refers to code token and AST node representation; SenRe refers to additional sentence-level retrieval. All of our results are the mean of 5 runs with different random seeds.}
\end{table*}

\paragraph{Metrics.}
Similar to recent work \citep{gong2022source, son_etal_2022_boosting}, we evaluate the source code summarization performance using three widely-used metrics, BLEU \citep{papineni}, METEOR \citep{banerjee_lavie_2005_meteor} and ROUGE-L \citep{lin_2004_rouge}. Furthermore, considering the essence of source code summarization to help humans better understand code, we also conduct a \textbf{human evaluation} study. The volunteers are asked to rank summaries generated from the anonymized approaches from 1 to 5 (i.e., 1: Poor, 2: Marginal, 3: Acceptable, 4: Good, 5: Excellent) based on \textbf{\textit{Similarity}}, \textbf{\textit{Relevance}}, and \textbf{\textit{Fluency}} metrics. Further details on human evaluation can be found in Appendix \ref{sec_human_appendix}.

\paragraph{Training Details.}
We implement our approach based on JoeyNMT \citep{kreutzer_etal_2019_joey}. The batch size is set to 32 and Adam optimizer is used with an initial learning rate $10^{-4}$. To alleviate overfitting, we adopt early stopping with patience 15.
For Faiss \citep{johnson2019billion} Index, we employ IndexFlatIP and top-$K$=16 to maintain a balance between retrieval quality and retrieval speed in the large-scale datastore. It is worth noting that only the base model requires training, and once trained, all the parameters of the base model are fixed.
For validation, we use greedy search, while for evaluation, we use beam search with beam size of 4. 

\subsection{Baselines}
\paragraph{Transformer-based.}
Transformer \citep{ahmad_etal_2020_transformer} is the first attempt to use transformer architecture in this field. Soon, structure-aware methods were proposed. Among these are CAST \citep{shi_etal_2021_cast} and mAST+GCN \citep{choi_etal_2021_learning}, which integrate structural information in a hybrid manner. SiT \citep{wu_etal_2021_code}, SiT+PDG \citep{son_etal_2022_boosting}, and CODESCRIBE \citep{guo-etal-2022-modelinghie} utilize a structured-guided way. The detailed description of these baselines is shown in Appendix \ref{sec_baseline}.

\paragraph{Retrieval-based.}
Rencos \citep{zhang2020retrieval} is the first retrieval-based Seq2Seq model, which computes a joint probability conditioned on both the original source code and the retrieved most similar source code for a summary generation. HGNN \citep{liu2021retrievalaugmented} is the retrieval-based GNN model, which retrieval the most similar code and uses a Hybrid GNN by fusing static graph and dynamic graph to capture global code graph information.

\begin{table*}
\centering
\resizebox{\textwidth}{!}{
\begin{tabular}{l|ccc|ccc}
\hline \hline
\multirow{2}{*}{\textbf{Model}} & \multicolumn{3}{c}{\textbf{CCSD}} & \multicolumn{3}{|c}{\textbf{Python$^\ddag$}} \\
\cline{2-7}
& \textbf{BLEU} & \textbf{ROUGE-L} & \textbf{METEOR} & \textbf{BLEU} & \textbf{ROUGE-L} & \textbf{METEOR} \\
\hline
\textit{Retrieval-based Methods} & & & & & & \\
Rencos \citep{zhang2020retrieval} & 14.80 & 31.41 & 14.64 & 34.73 & 47.53 & 21.06 \\
HGNN \citep{liu2021retrievalaugmented} & 16.72 & 34.29 & 16.25 & - & - & - \\
\hline 
\textit{Our Method} & & & & & &\\
\hspace{1em}Base   & 17.82 & 35.33 & 16.71 &  34.85 & 48.84 & 21.49 \\
\hspace{1em}Base + Rencos & 19.43 & 36.92 & 17.69 & 35.26 & 49.25 & 22.07 \\
\hspace{1em}Tram w/o HR & 21.27 & 37.61 & 18.09 & 36.41 & 50.18 & 22.24 \\
\hspace{1em}Tram & \textbf{21.48} & \textbf{37.88} & \textbf{18.35} & \textbf{36.73} & \textbf{50.35} & \textbf{22.53} \\
\hspace{1em}\textit{Tram with SenRe} & \textit{22.23} & \textit{38.16} & \textit{18.96} & \textit{36.95} & \textit{50.69} & \textit{22.93} \\
\hline \hline 
\end{tabular}
}
\caption{\label{main_result_ccsd}
Comparison of other retrieval methods. 
HR means code token and AST node representation; SenRe means additional sentence-level retrieval. All of our results are the mean of 5 runs with different random seeds.
}
\end{table*}

\begin{table*}
\small
\centering
\begin{tabular}{l|ccc|ccc}
\hline 
\multirow{2}{*}{\textbf{Model}} & \multicolumn{3}{c|}{\textbf{Java}} & \multicolumn{3}{c}{\textbf{Python$^\ddag$}} \\
\cline{2-7}
& \textbf{Similarity} & \textbf{Relevance} & \textbf{Fluency} & \textbf{Similarity} & \textbf{Relevance} & \textbf{Fluency} \\
\hline
Rencos  & - & - & - & 3.07 & 3.06 & 3.96 \\
CODESCRIBE   & 3.67 & 3.72 & 4.16 & - & - & -\\
Base & 3.62 & 3.64 & 4.10 & 3.20 & 3.24 & 4.03\\
Tram & \textbf{3.83} & \textbf{3.89} & \textbf{4.23} & \textbf{3.33} & \textbf{3.44} & \textbf{4.14} \\
\hline 
\end{tabular}
\caption{\label{human}
Human Evaluation on Java and Python$^\ddag$ datasets.
}
\end{table*}

\subsection{Main Results}
The main experiment results are shown in Table \ref{main_result_java_python} and Table \ref{main_result_ccsd} in terms of three automatic evaluation metrics. The reason we have two tables is that transformer-based works compare their performance on the widely-used Java and Python benchmarks, while the retrieval-based works use two different benchmarks, namely CCSD and Python$^\ddag$. Thus, our experiments are performed on all four datasets for a more thorough comparison. We calculate the metric values following the same scripts\footnote{\url{https://github.com/gingasan/sit3/blob/main/c2nl/eval/bleu/google_bleu.py}}. 

From Table \ref{main_result_java_python}, SiT + PDG and CODESCRIBE achieve better results than all previous works. However, it is worth noting that even our base model can achieve comparable performance to other models. This is due to the improved training method we used, Pre-LN (layer normalization inside the residual blocks), which is discussed in \citep{liu-etal-2020-understanding}. This method enhances the stability of the training process and leads to better performance. Tram further boosts results with 1.39 BLEU points on Java and 1.53 BLEU points on Python and achieves new state-of-the-art results. We also observe that the performance improvement for Python is better than that for Java. The main reason we speculate is that Java has a longer average code token length (from Table \ref{tab: statistics}) and richer code structure information.

In Table \ref{main_result_ccsd}, we compare Tram with other retrieval-based models on CCSD and Python$^\ddag$ benchmarks. Our base model is even superior to other retrieval-based methods; the main reason is that the backbone \footnote{Other retrieval-based methods are RNN-based.} are different. We reproduce Rencos architecture\footnote{HGNN code is not open source.} in our base model for a fair comparison, which we denoted as ``Base + Rencos''. Tram outperforms all other retrieval-based methods, further improving performance with 2.05 BLEU points and 1.47 BLEU points on CCSD and Python$^\ddag$, respectively. Furthermore, as shown in Table \ref{main_result_java_python} and \ref{main_result_ccsd}, enhancing Tram with additional sentence-level retrieval (refer as \textit{"Tram with SenRe"}) and its integration with code pre-trained models (\textit{"Our Method on Pre-trained Models"} section in Table \ref{main_result_java_python}) leads to a notable improvement in performance.

\begin{table*}[ht]
\small
\centering
\begin{tabular}{l}
\hline
\begin{lstlisting}[   % 进行参数设置
 language=C, % 设置语言
 basicstyle=\small,
 basicstyle=\ttfamily\footnotesize, % 设置字体族
 breaklines=true, % 自动换行
 keywordstyle=\bfseries\color{NavyBlue}, % 设置关键字为粗体，颜色为 NavyBlue
 morekeywords={}, % 设置更多的关键字，用逗号分隔
 emph={cfg}, % 指定强调词，如果有多个，用逗号隔开
    emphstyle=\bfseries\color{Rhodamine}, % 强调词样式设置
    commentstyle=\itshape\color{black!50!white}, % 设置注释样式，斜体，浅灰色
    stringstyle=\bfseries\color{PineGreen!90!black}, % 设置字符串样式
    columns=flexible,
    showstringspaces=false,
]
void scsi_netlink_init(void){
    struct netlink_kernle_cfg cfg;
    cfg.input  = scsi_nl_rcv_msg;
    cfg.groups = SCSI_NL_GPRP_CNT;
    scsi_nl_sock = netlink_kernel_create(&init_net,
    NETLINK_SCSITRANSPORT, &cfg);
    if (!scsi_nl_sock){
        printk(KERN_ERR "%s: register of receive handler failed\n", __func__);
        return;}
    return;}
\end{lstlisting} \\
\hline
\textcolor[RGB]{0,0,255}{Base:} called by scsi netlink initialization to register the scsi netlink interface. \\
\textcolor[RGB]{60,179,113}{Rencos:} called by scsi netlink interface to register the scsi netlink interface. \\
\textcolor[RGB]{255,140,0}{Tram:} called by scsi \textcolor[RGB]{255,0,0}{\textbf{subsystem}} to register the scsi transport netlink interface. \\
\textcolor[RGB]{205,50,120}{Human Written:} called by scsi \textcolor[RGB]{255,0,0}{\textbf{subsystem}} to initialize the scsi transport netlink interface. \\
\textcolor[RGB]{255,0,0}{\textit{\textbf{Retrieval Results:}}} ``\textit{subsystem}'' (0.90), ``\textit{transport}''(0.04), \textit{``stack''}(0.02), \textit{``command''}(0.0034), \textit{``device''}(0.0025) $\cdots$\\
\hline 
\end{tabular}
\caption{A Python instance. The bold red font is the keyword of generated summary. The \textbf{\textit{Retrieval Results}} line is the visible retrieval results and corresponding probability after applying $softmax$ on the keyword generation step.}
\label{tab:quality-main}
\end{table*}

\subsection{Ablation Study}
To validate the effectiveness of intelligently fusing summary token representation with code token representation $H_t$ and AST node representation $R_t$, we conduct an ablation experiment where we eliminate the $H_t$, $R_t$, and directly use $d_{t-1}$ to represent target summary token $s_t$ for comparison (refer as ``Tram w/o HR''). As shown in Table \ref{main_result_java_python} and \ref{main_result_ccsd}, the performance declined by 0.47, 0.60, 0.21, and 0.32 BLEU points for Java, Python, CCSD, and Python$^\ddag$, respectively. This decline in performance across all datasets demonstrated the importance of fusing code semantics into the summary token for effective token-level retrieval on the decoder side.

\begin{figure}[ht]
\centering  
\subfigure{
\label{Fig.sub.1}
\includegraphics[width=0.48\linewidth]{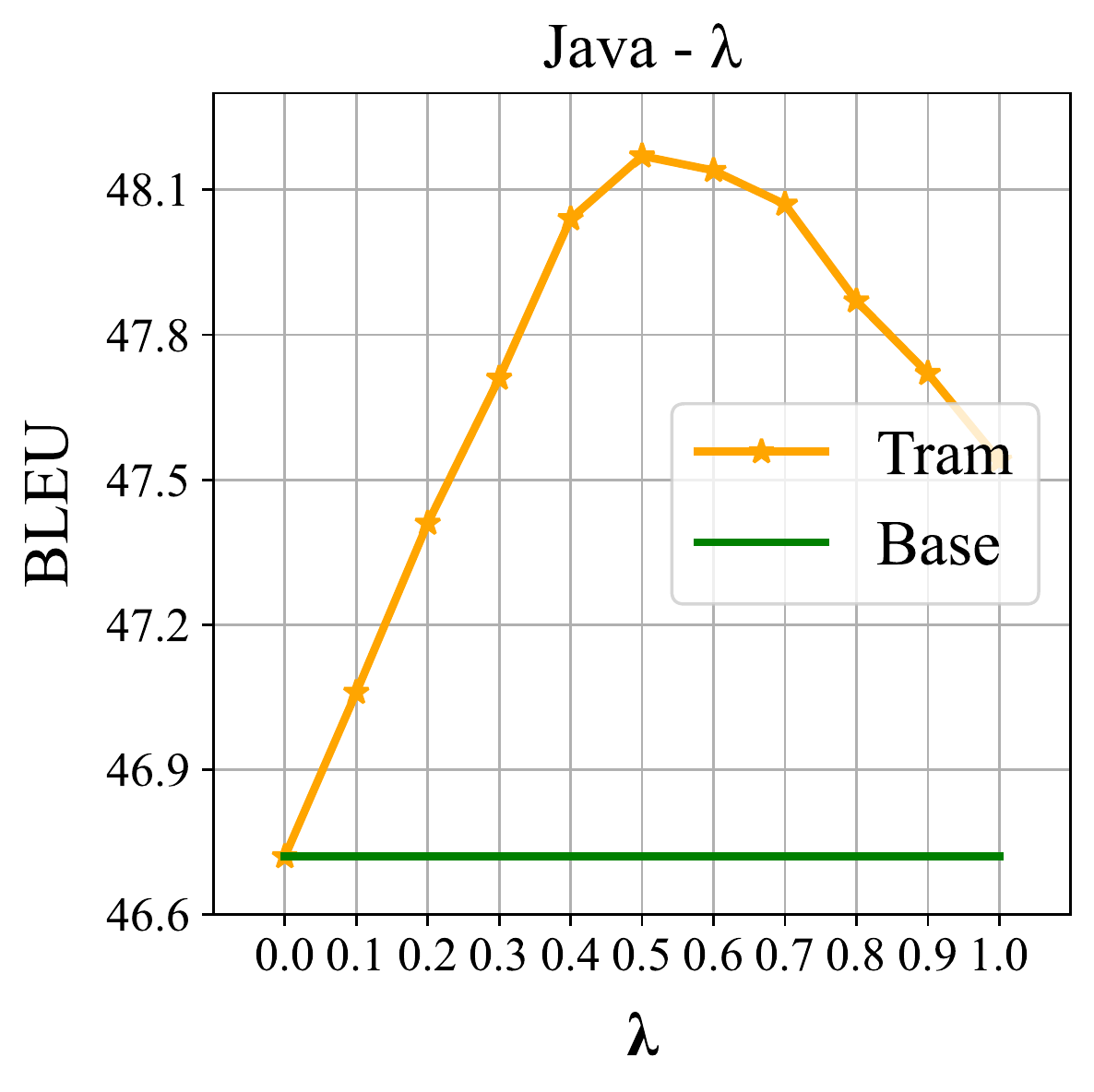}}
\subfigure{
\label{Fig.sub.2}
\includegraphics[width=0.48\linewidth]{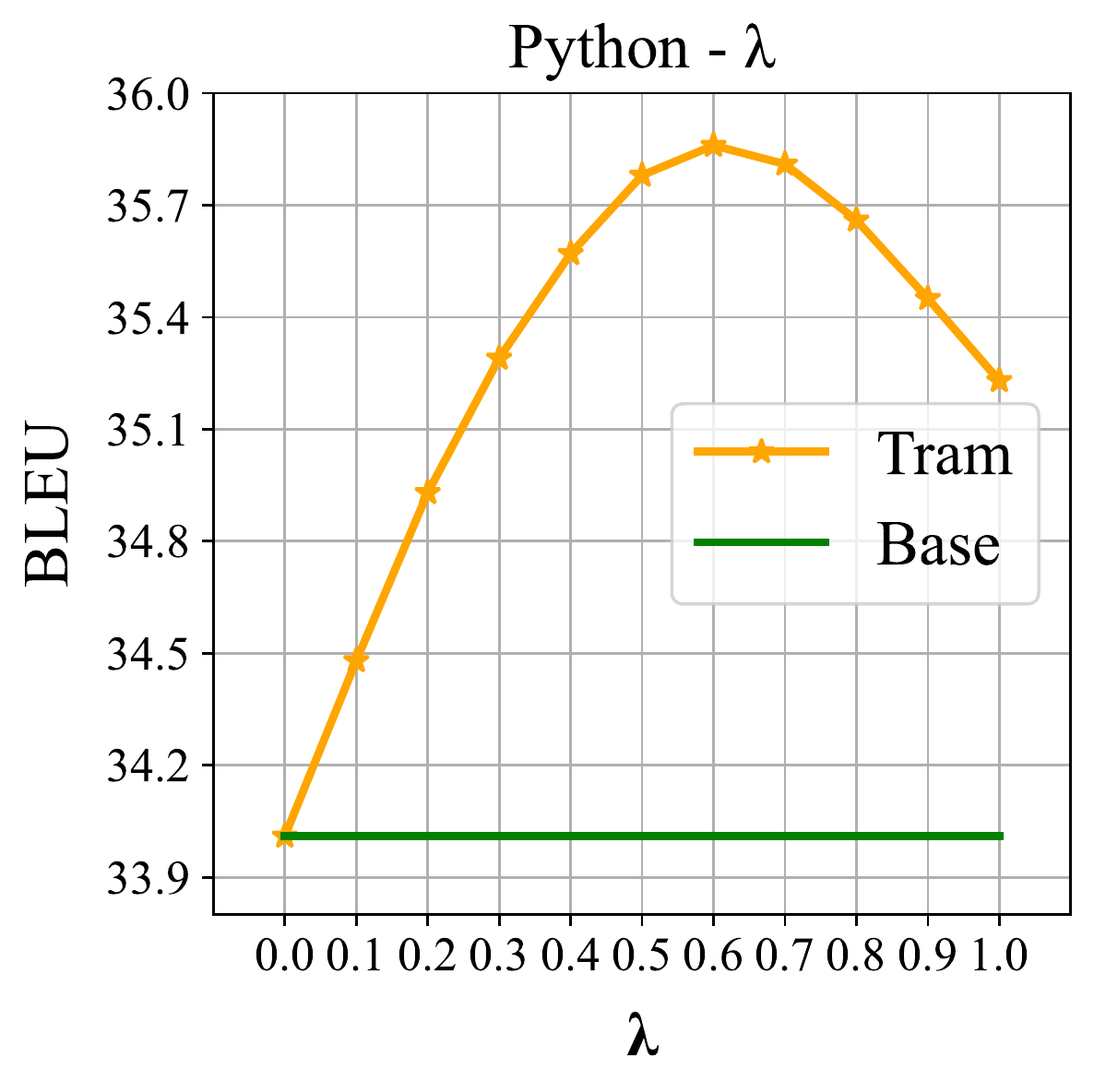}}
\subfigure{
\label{Fig.sub.3}
\includegraphics[width=0.48\linewidth]{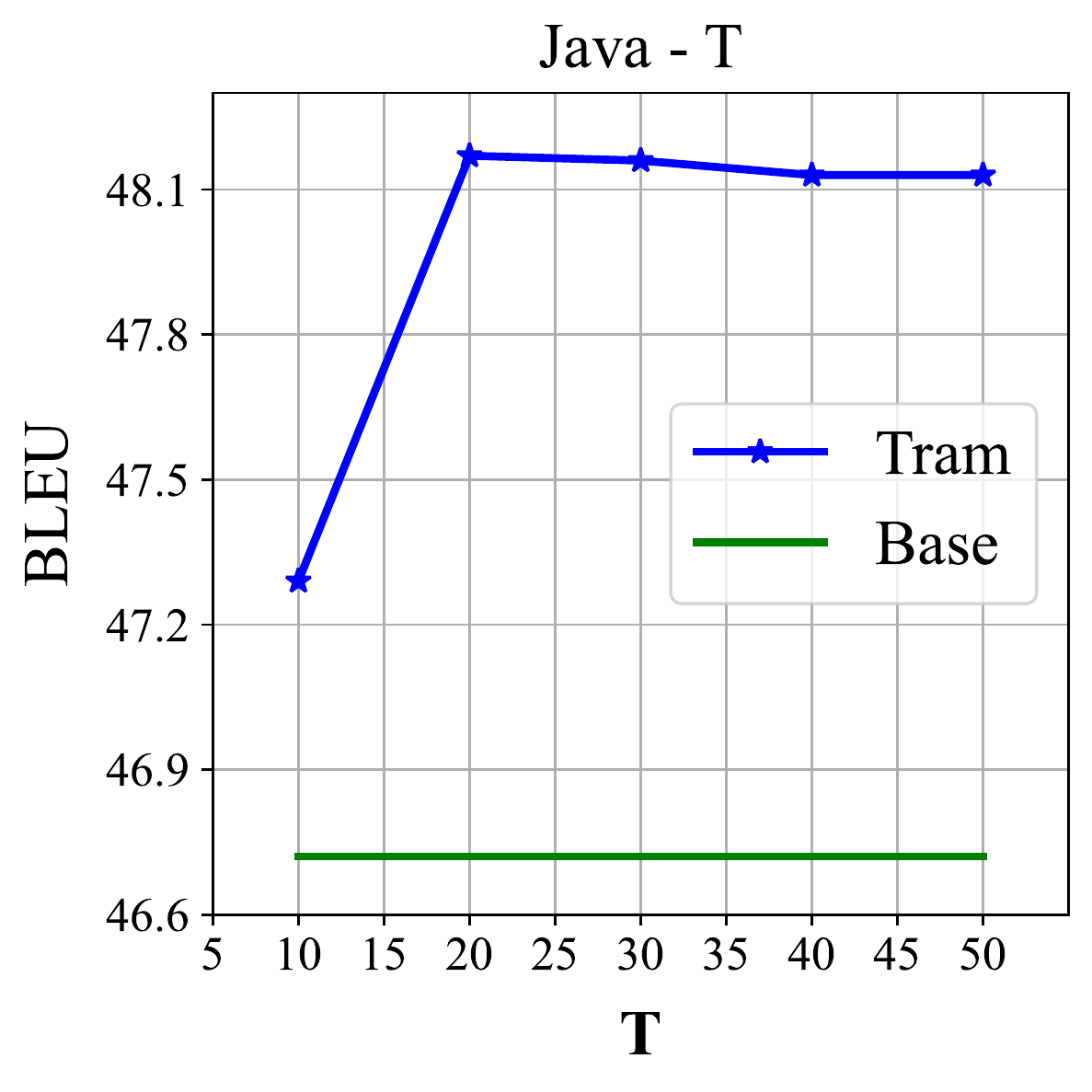}}
\subfigure{
\label{Fig.sub.4}
\includegraphics[width=0.48\linewidth]{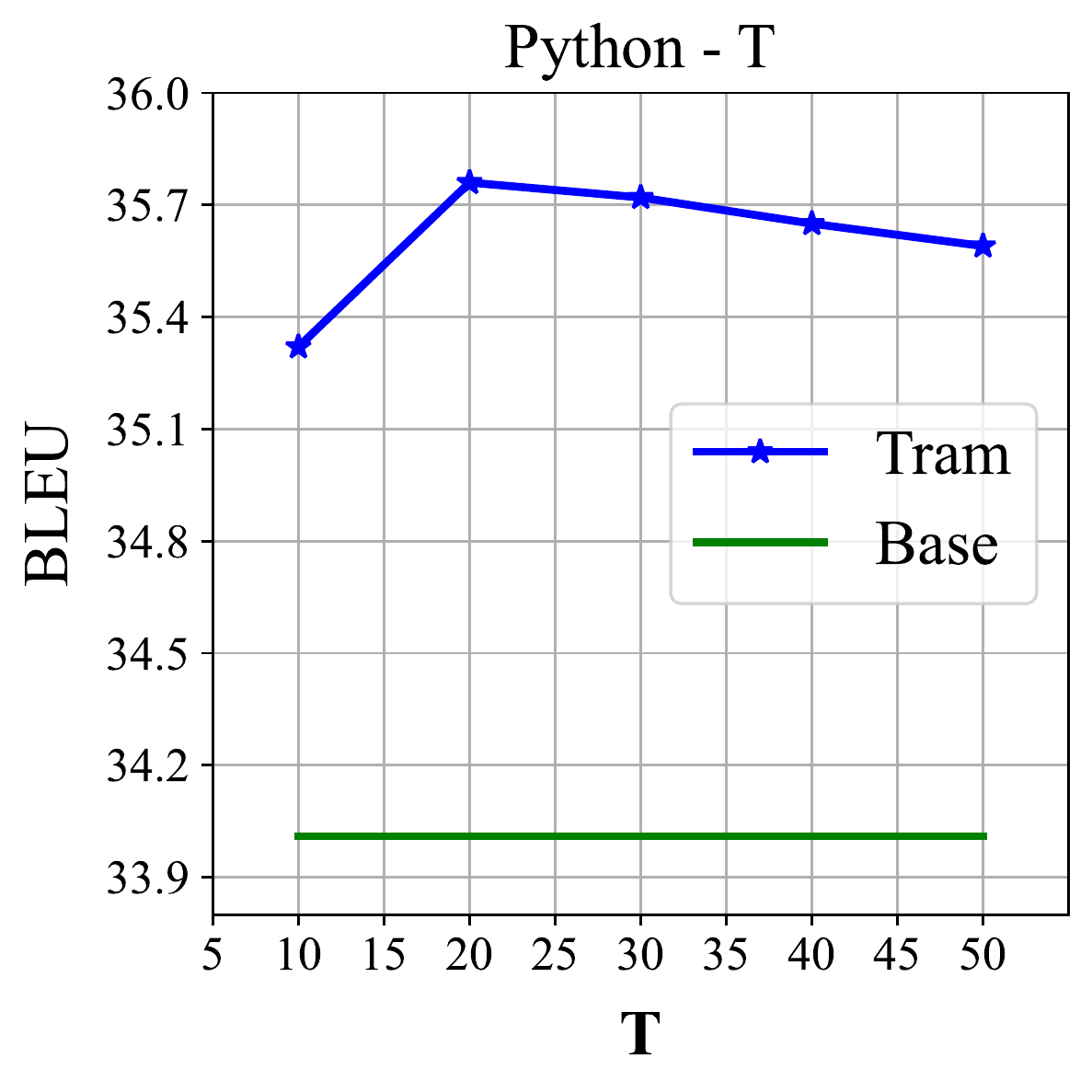}}
\caption{$\lambda$ and $T$ selections in Java and Python datasets. 
}
\label{fig:hyperparameters}
\end{figure}

\subsection{Human Evaluation}
We perform a human evaluation (details provided in Appendix \ref{sec_human_appendix}) to assess the quality of the generated summaries by Tram, Rencos, CODESCRIBE, and base model in terms of \textit{Similarity}, \textit{Relevance}, and \textit{Fluency} as shown in Table \ref{human}. 
The results show that Tram can generate better summaries that are more similar to the ground truth, more relevant to the source code, and more fluent in naturalness.

\section{Analysis}
\subsection{Hyperparameters Analysis}
Tram has two primary hyperparameters: $\lambda$ and $T$. $\lambda$ means the weight of the retrieval-based distribution component in the final distribution; the higher value indicates greater reliance on retrieval results, and vice versa. $T$ means temperature, which smooths the retrieval-based distribution. We plot the performance of Tram with different hyperparameter selections in Figure \ref{fig:hyperparameters}. The value of $\lambda$ has a significant impact on the final performance, and we find that different datasets have different optimal values (i.e., $\lambda=0.5$ for Java and $\lambda=0.6$ for Python). We also observe that $\lambda$ = 1 outperforms $\lambda$ = 0. The reason is related to the BLEU score (detailed cause analysis provided in Appendix \ref{sec_lambda}). Regarding $T$, if it is too small, the retrieval-based distribution cannot be adequately distinguished; while if it is too large, the retrieval-based distribution will concentrate on a single token. Our final results indicate that both extremes result in a performance decrease.

\begin{table}[ht]
\centering
\resizebox{0.5\textwidth}{!}{
\begin{tabular}{ll|cccccc}
\hline
\multicolumn{2}{l|}{Token Frequency}                 & 1   & 2  & 5  & 10 & 50 & 100 \\ \hline
\multirow{3}{*}{Java}                       & Base   & 126 & 75 & 45 & 27 & 28 & 16  \\
                                            & Rencos & 243 & 138 & 73 & 38 & 37 & 18  \\
                                            & Tram   & 307 & 164 & 115 & 51 & 42 & 21  \\ \hline
\multicolumn{1}{c}{\multirow{3}{*}{Python$^\ddag$}} & Base   & 452 & 376 & 272 & 176 & 84 & 82  \\
\multicolumn{1}{c}{}                        & Rencos & 799 & 515 & 344 & 223 & 88 & 109  \\
\multicolumn{1}{c}{}                        & Tram   & 983 & 647 & 405 & 298 & 103 & 121  \\ \hline
\end{tabular}
}
\caption{Count of Accurately Generated Low-Frequency Tokens.}
\label{tab:frequency}
\end{table}

\subsection{Token Frequency In-Depth Analysis}
Compared to the coarse-grained retrieval approach at the sentence-level, the token-level retrieval can capture the top-$K$ most semantically relevant tokens at every step. 
This can increase the likelihood of generating those low-frequency tokens in the summary text. Since these low-frequency tokens and their corresponding representations are stored in the datastore, by retrieving the most semantically similar tokens at each generation step, these low-frequency tokens can be more easily and directly fetched from the datastore compared to purely model generated. We further conduct an in-depth statistical analysis of the generation quantity of low-frequency tokens. We first collect all the correctly generated tokens according to the ground-truth summaries. Then we count the frequencies of all these correct tokens in the training set and record the number of the correct and low-frequency tokens (frequency = 1, 2, 5, 10, 50, 100). From Table \ref{tab:frequency}, we can see that Tram can correctly predict more low-frequency tokens than Rencos (sentence-level retrieval) and Base (vanilla model generated) when the token frequency is small ($\leq 100$).

\begin{table}[ht]
\centering
\resizebox{0.48\textwidth}{!}{
\begin{tabular}{ccccc}
\hline 
        \textbf{Python}     & \textbf{Datastore}     & \textbf{BLEU}   & \textbf{ROUGE-L}  & \textbf{METEOR}   \\ \hline 
                      & Vanilla        & 35.97 & 49.92 & 22.09  \\
                      & Noise-5\%    & 35.84 & 49.79 & 21.98  \\
                      & Noise-10\%   & 35.68 & 49.67 & 21.85   \\
                      & Noise-20\%   & 35.49 & 49.33 & 21.70   \\ \hline 
             \textbf{Java}       & \textbf{Datastore} & \textbf{BLEU}   & \textbf{ROUGE-L}  & \textbf{METEOR}   \\ \hline 
                      & Vanilla   & 48.32 & 58.13 & 29.56   \\
                      & Noise-5\%    & 48.15 & 57.95 & 29.44 \\
                       & Noise-10\%   & 48.07 & 57.90 & 29.37  \\
                      & Noise-20\%   & 47.82 & 57.61 & 28.81  \\
\hline 
\end{tabular}
}
\caption{Datastore Quality and Robustness
Analysis at Different Noise Levels.}
\label{tab:robustness}
\end{table}

\subsection{Datastore Quality and Robustness Analysis}
To accurately assess the impact of datastore quality on Tram's performance, we conduct robustness experiments where noise is intentionally introduced into the datastore. Specifically, we randomly shuffle a certain percentage of (representation, token) pairs, leading to misaligned pairings. These experiments, conducted using Python and Java datasets, are based on the averages from five separate runs. We introduce noise levels of 5\%, 10\%, and 20\%, corresponding to the proportion of misaligned pairs in the datastore. Table \ref{tab:robustness} presents the experimental results, indicating that even with a 10\% noise level in the datastore, the BLEU score reduction is only up to 0.3 points. Furthermore, even under 20\% noise conditions, the model maintains robust performance. These results suggest that the impact of datastore quality and the presence of noisy or poorly aligned pairs is relatively minimal, confirming the robustness of both the datastore and our Tram method.

\subsection{Qualitative Analysis}
We provide a python example in Table \ref{tab:quality-main} to demonstrate the effectiveness and interpretability of Tram. The qualitative analysis reveals that, compared to other models, Tram enables visualization of the \textit{Retrieval Results} and corresponding probability at each generation step, as depicted in the last line, making our approach more interpretable. More visualized instances can be found in Appendix \ref{sec_examples}.


\section{Conclusion}
In this paper, we propose a novel token-level retrieval-augmented mechanism for source code summarization. By a well-designed fine-grained retrieval pattern, Tram can effectively incorporate external human-written code-summary pairs on the decoder side. Extensive experiments and human evaluation show that Tram not only significantly improves performance but also generates more low-frequency tokens and enhances interpretability. 


\section*{Limitations}
Our retrieval-augmented method (Tram) takes full advantage of external retrieval information, and the performance improvement relies on high-quality code-summary token-level pairs. However, there exists some noise in the datastore which will bias the final token distribution; therefore, dealing with noise deserves our deeper exploration. Furthermore, our experiments are only on high-resource programming language (Python, Java, C) scenarios; exploring how to apply our model in a low-resource programming language (Ruby, Go, etc.) is our future direction.

\section*{Acknowledgements}
This work was partly supported by NSFC under Grant No. 62302443, the Fellowship of China National Postdoctoral Program for Innovative Talents (BX20230307), the Fundamental Research Funds for the Central Universities (Zhejiang University NGICS Platform). This research was also supported by the advanced computing resources provided by the Supercomputing Center of Hangzhou City University.


\bibliography{anthology,custom}

\appendix

\section{Integration of Tram with Code Pre-trained Models}
\label{sec_tram_to_llms}
We need to clarify that our Tram can be integrated with generative code pre-trained models (encoder-decoder architecture), such as CodeT5 \citep{wang-etal-2021-codet5} and UniXcoder \citep{guo-etal-2022-unixcoder}, but is not suitable for code pre-trained models used for code understanding (encoder-only architecture), like CodeBERT \citep{feng-etal-2020-codebert} and GraphCodeBERT \citep{guo2021graphcodebert}.

Specifically, the integration process is similar to the \nameref{methodology} section and primarily consists of three steps:

(1) We use Java \citep{hu2018deep} and Python \citep{wan2018improving} datasets to fine-tune the code pre-trained models, respectively, and treat the fine-tuned models as base models;

(2) During the datastore establishment phase, the process aligns with that described in the \nameref{datastore_construction} section. However, we have omitted the AST input to satisfy the input conditions of the code pre-trained models;

(3) Token-level Retrieval: The retrieved top-$K$ tokens are expanded to a probability distribution (which we refer to as the retrieval-based distribution). Then we fused the retrieval-based distribution with the vanilla distribution built on the original vocabulary table of the code pre-trained models to obtain the final distribution.

\section{Details on Transformer-based Methods}
\label{sec_baseline}
Transformer \citep{ahmad_etal_2020_transformer} is the first attempt to use transformer architecture, equipped with relative positional encoding and copy mechanism \citep{see_etal_2017_get}, effectively capturing long-range dependencies of source code. CAST \citep{shi_etal_2021_cast} hierarchically splits a large AST into a set of subtrees and utilizes a recursive neural network to encode the subtrees. The aim is to capture the rich information in ASTs. mAST + GCN \citep{choi_etal_2021_learning} adopt the AST and graph convolution to model the structural information and the transformer to model the sequential information. SiT \citep{wu_etal_2021_code} incorporates a multi-view graph matrix into the transformer's self-attention mechanism. 
SiT + PDG \citep{son_etal_2022_boosting} points program dependency graph is more effective for expressing the structural information than AST.
CODESCRIBE \citep{guo-etal-2022-modelinghie} model the hierarchical syntax structure of code by introducing a novel triplet position.

\section{Human Evaluation}
\label{sec_human_appendix}
In our human evaluation, we invited 3 PhD students and 5 master students with at least 2-5 years of software engineering experience as volunteers. We conduct a small-scale random dataset (i.e., 100 random Java samples and 100 random Python samples). 
 The volunteers are asked to rank summaries generated from the anonymized approaches from 1 to 5 (i.e., 1: Poor, 2: Marginal, 3: Acceptable, 4: Good, 5: Excellent) based on the three following questions:
\begin{itemize}
    \item \textbf{Similarity}: How similar of generated summary and ground truth?
    \item \textbf{Relevance}: Is the generated summary relevant to the source code?
    \item \textbf{Fluency}: Is the generated summary syntactically correct and fluent?
\end{itemize}
For each evaluation summary, the rating scale is from 1 to 5, where a higher score means better quality. Responses from all volunteers are collected and averaged.

\section{Cause Analysis: Performance Superiority of $\lambda = 1$ over $\lambda = 0$}
\label{sec_lambda}
 $\lambda$ means the weight of the retrieval-based distribution component in the final distribution. The reason is related to the BLEU score. The BLEU metric measures the similarity between two sentences by assessing the overlap of words between them. Model-generated sentences tend to produce more common words, leading to better fluency; in contrast, sentences generated through retrieval methods are more likely to include factual terms, which, when evaluated using the BLEU score, results in a higher score \citep{reiter-2018-structured}. However, it may scarify the language quality. 
 
 For example, given the ground truth \textbf{"start a source file within a compilation unit."}, the retrieval-based generation with $\lambda = 1$: \textbf{"start file within a compilation unit unit."}, achieves a BLEU score of 48.78. This is higher than the model-based generation with $\lambda = 0$: \textbf{"start the source file within the unit."}, which scores a BLEU of 33.17. Indeed, neither $\lambda = 1$ or $\lambda =  0$ is good enough, and we need a trade-off between the retrieval and the model generation.

\section{Qualitative Examples}
\label{sec_examples}
Table \ref{tab:quality-appendix} shows a couple of qualitative examples to demonstrate the effectiveness and interpretability of Tram.
\begin{table*}[ht]
\small
\centering
\begin{tabular}{l}
\hline
\begin{lstlisting}[   % 进行参数设置
 language=C, % 设置语言
 basicstyle=\small,
 basicstyle=\ttfamily\footnotesize, % 设置字体族
 breaklines=true, % 自动换行
 keywordstyle=\bfseries\color{NavyBlue}, % 设置关键字为粗体，颜色为 NavyBlue
 morekeywords={}, % 设置更多的关键字，用逗号分隔
 emph={bat_priv, bat_attr}, % 指定强调词，如果有多个，用逗号隔开
    emphstyle=\bfseries\color{Rhodamine}, % 强调词样式设置
    commentstyle=\itshape\color{black!50!white}, % 设置注释样式，斜体，浅灰色
    stringstyle=\bfseries\color{PineGreen!90!black}, % 设置字符串样式
    columns=flexible,
    showstringspaces=false,
]
void batadv_sysfs_del_meshif(struct net_device *dev)
{
    struct batadv_priv *bat_priv = netdev_priv(dev);
    struct batadv_attribute **bat_attr;
    for (bat_attr = batadv_mesh_attrs; *bat_attr; ++bat_attr)
        sysfs_remove_file(bat_priv->mesh_obj, &((*bat_attr)->attr));

    kobject_uevent(bat_priv->mesh_obj, KOBJ_REMOVE);
    kobject_del(bat_priv->mesh_obj);
    kobject_put(bat_priv->mesh_obj);
    bat_priv->mesh_ojb = NULL;
}
\end{lstlisting} \\
\hline
\textcolor[RGB]{0,0,255}{Base:} Remove mesh interface-related sysfs sysfs entries. \\
\textcolor[RGB]{60,179,113}{Rencos:} Delete mesh junction sysfc attributes. \\
\textcolor[RGB]{255,140,0}{Tram:} Remove soft \textcolor[RGB]{255,0,0}{\textbf{interface}} specific sysfs entries. \\
\textcolor[RGB]{205,50,120}{Human Written:} Remove soft \textcolor[RGB]{255,0,0}{\textbf{interface}} specific sysfs entries. \\
\textcolor[RGB]{255,0,0}{\textit{\textbf{Retrieval Results:}}} ``\textit{interface}'' (0.82), ``\textit{portal}''(0.11), \textit{``bridge''}(0.04), \textit{``junction''}(0.0086), \textit{``link''}(0.0013) $\cdots$\\
\hline 
\begin{lstlisting}[   % 进行参数设置
 language=Python, % 设置语言
 basicstyle=\small,
 basicstyle=\ttfamily\footnotesize, % 设置字体族
 breaklines=true, % 自动换行
 keywordstyle=\bfseries\color{NavyBlue}, % 设置关键字为粗体，颜色为 NavyBlue
 morekeywords={}, % 设置更多的关键字，用逗号分隔
 emph={site, reverse}, % 指定强调词，如果有多个，用逗号隔开
    emphstyle=\bfseries\color{Rhodamine}, % 强调词样式设置
    commentstyle=\itshape\color{black!50!white}, % 设置注释样式，斜体，浅灰色
    stringstyle=\bfseries\color{PineGreen!90!black}, % 设置字符串样式
    columns=flexible,
    % frame={topline, bottomline} % , rightline
]
def category_structure(category, site): 
    return {'description': category.title, 
    'html_Url': ('%s://%s%s'%(PROTOCOL, site.domain, 
                category.get_absolute_url())), 
    'rss_Url': ('%s://%s%s'%(PROTOCOL, site.domain, 
                reverse('zinnia:category_feed', args=[category.tree_path]))), 
    'category_Id': category.pk , 
    'parent_Id': ((category.parent and category.parent.pk) or 0 ), 
    'category_Description': category.description, 
    'category_Name': category.title } 
\end{lstlisting} \\
\hline
\textcolor[RGB]{0,0,255}{Base:} updates the structure. \\
\textcolor[RGB]{60,179,113}{Rencos:} a post structure. \\
\textcolor[RGB]{255,140,0}{Tram:} a \textcolor[RGB]{255,0,0}{\textbf{category}} structure. \\
\textcolor[RGB]{205,50,120}{Human Written:} a \textcolor[RGB]{255,0,0}{\textbf{category}} structure. \\
\textcolor[RGB]{255,0,0}{\textit{\textbf{Retrieval Results:}}} ``\textit{category}''(0.43), ``\textit{tag}''(0.11), ``\textit{post}''(0.07), ``\textit{helper}''(0.06), ``\textit{version}''(0.06) $\cdots$ \\
\hline
\end{tabular}
\caption{Task samples. The first is a C instance; the second is a Python instance. The bold red font is the keyword of the generated summary. The \textbf{\textit{Retrieval Results}} line is the visible retrieval results and corresponding probability after applying $softmax$ on the keyword generation step.}
\label{tab:quality-appendix}
\end{table*}

\end{document}